\documentclass{article}
\PassOptionsToPackage{numbers, compress}{natbib}

\usepackage[final]{neurips_2024}

\usepackage[utf8]{inputenc} 
\usepackage[T1]{fontenc}    
\usepackage{hyperref}       
\usepackage{url}            
\usepackage{booktabs}       
\usepackage{amsfonts}       
\usepackage{nicefrac}       
\usepackage{microtype}      
\usepackage{xcolor}         
\usepackage{graphbox}
\usepackage{multirow}
\usepackage{makecell}
\usepackage{wrapfig}

\newcommand*\samethanks[1][\value{footnote}]{\footnotemark[#1]}

\usepackage{amsmath,amsfonts,bm,amssymb,mathtools,amsthm}
\usepackage{color,xcolor,xspace}
\usepackage{booktabs}
\usepackage{thm-restate}

\def\eqref#1{Eq.~(\ref{#1})}

\def\1{\bm{1}}

\def\rmP{{\mathbf{P}}}

\def\vm{{\bm{m}}}
\def\vn{{\bm{n}}}

\def\vx{{\bm{x}}}

\def\vz{{\bm{z}}}

\DeclareMathAlphabet{\mathsfit}{\encodingdefault}{\sfdefault}{m}{sl}
\SetMathAlphabet{\mathsfit}{bold}{\encodingdefault}{\sfdefault}{bx}{n}

\def\gH{{\mathcal{H}}}

\def\gL{{\mathcal{L}}}

\newcommand{\E}{\mathbb{E}}

\newcommand{\R}{\mathbb{R}}

\DeclareMathOperator*{\argmax}{arg\,max}

\usepackage[capitalise]{cleveref}

\title{Unlocking the Capabilities of Masked Generative Models for Image Synthesis via Self-Guidance}
\vspace{-5pt}
\author{
    \vspace{-25pt} \\
    \textbf{Jiwan Hur$^1$ \quad\quad Dong-Jae Lee$^1$ \quad\quad Gyojin Han$^1$ \quad\quad Jaehyun Choi$^1$}\vspace{3pt} \\  
    \textbf{Yunho Jeon$^2$\thanks{Corresponding authors: \href{mailto:junmo.kim@kaist.ac.kr}{\color{black}{junmo.kim@kaist.ac.kr}}, \href{mailto:yhjeon@hanbat.ac.kr}{\color{black}{yhjeon@hanbat.ac.kr}}} \quad\quad Junmo Kim$^1$\samethanks[2]}\vspace{4pt} \\
    $^1$KAIST, South Korea ~~\quad\quad $^2$Hanbat National University, South Korea \vspace{3pt} \\
    {\texttt{\small \{jiwan.hur, jhtwosun, hangj0820, chlwogus\}@kaist.ac.kr}} \\
    {\texttt{\small yhjeon@hanbat.ac.kr, junmo.kim@kaist.ac.kr}} \vspace{8pt} \\
    Code is available at: \url{https://github.com/JiwanHur/UnlockMGM} \vspace{-3pt}
}

\begin{document}

\maketitle

\begin{figure}[!h]
  \centering
  \vspace{-5mm}
\includegraphics[width=\linewidth]{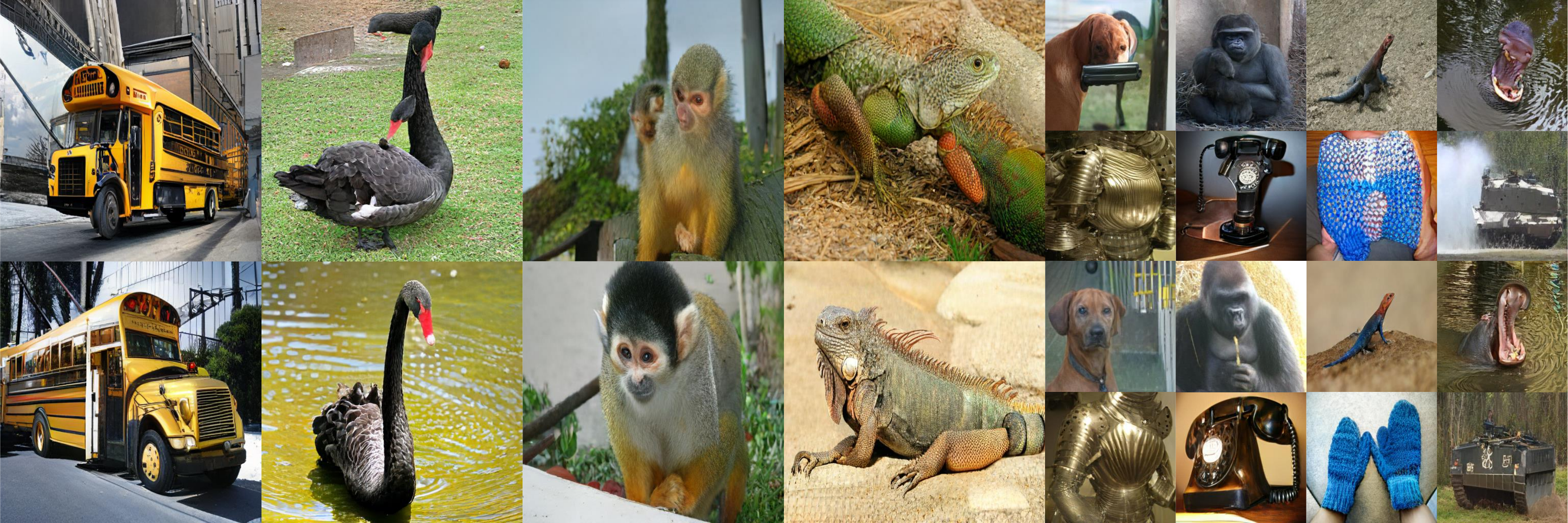}  
  \vspace{-5mm}
  \caption{
  \textbf{Comparison of sampled images using 18-step MaskGIT~\cite{maskgit} without (top) and with the proposed self-guidance (bottom) on ImageNet 512$\times$512 (left) and 256$\times$256 (right) resolutions.} 
  Each paired image is sampled using the same random seed and sampling hyperparameters. 
  The proposed self-guidance effectively improves the capabilities of the masked generative models. 
  }
  \label{fig:visualize}
  \vspace{-1mm}
\end{figure}

\begin{abstract}

Masked generative models (MGMs) have shown impressive generative ability while providing an order of magnitude efficient sampling steps compared to continuous diffusion models.
However, MGMs still underperform in image synthesis compared to recent well-developed continuous diffusion models with similar size in terms of quality and diversity of generated samples.
A key factor in the performance of continuous diffusion models stems from the guidance methods, which enhance the sample quality at the expense of diversity.
In this paper, we extend these guidance methods to generalized guidance formulation for MGMs and propose a self-guidance sampling method, which leads to better generation quality.
The proposed approach leverages an auxiliary task for semantic smoothing in vector-quantized token space, analogous to the Gaussian blur in continuous pixel space. 
Equipped with the parameter-efficient fine-tuning method and high-temperature sampling, MGMs with the proposed self-guidance achieve a superior quality-diversity trade-off, outperforming existing sampling methods in MGMs with more efficient training and sampling costs.
Extensive experiments with the various sampling hyperparameters confirm the effectiveness of the proposed self-guidance. 

\end{abstract}

\subsection*{Keywords}
Image synthesis, discrete diffusion models, masked generative models, sampling guidance, parameter-efficient fine-tuning

\setcounter{footnote}{0} 
\section{Introduction} \vspace{-3mm}

With the advent of generative adversarial networks (GANs)~\cite{GAN}, generative models have attracted significant attention for their powerful ability to synthesize highly realistic images.
However, due to the limited mode coverage and training instability, recently, likelihood-based models such as diffusion models~\cite{ho2020denoising} have been actively researched.
Diffusion models have shown promising results for their diverse and high-quality samples, surpassing GANs on class conditional image synthesis~\cite{adm}.

A key factor in the success of diffusion models stems from the various guidance techniques, which enhance the fidelity of the generated image at the expense of diversity. 
The guidance is usually conducted throughout the sampling process of diffusion models, driving it toward enhancing specific information such as class~\cite{adm,ho2022classifier}, text~\cite{ldm,improvedvq}, or details in the image~\cite{sag}.
From the practical perspective, however, diffusion models suffer from sampling inefficiency, requiring hundreds of sampling steps to generate high-quality images.
Moreover, guidance techniques further decrease the sampling efficiency, as they typically require twice as many model inferences as the original process. For instance, ADM~\cite{adm} with guidance requires $\sim$500 model inferences.

On the other side, masked generative models (MGMs) have shown superior trade-offs between sampling quality and speed compared to (continuous) diffusion models~\cite{maskgit,tokencritic,dpc}.
MGMs use an absorbing state (\texttt{[MASK]}) diffusion process~\cite{d3pm} and aim to generate discrete tokens by predicting the masked region, similar to BERT~\cite{devlin2018bert}. Specifically, they use the Markov transition matrix to model the diffusion process and sample the tokens with the categorical distribution.
Recently, vector-quantized (VQ) image token-based MGMs with (non-)autoregressive transformer~\cite{esser2021taming, maskgit} have demonstrated an efficient sampling process, providing an order of magnitude fewer sampling steps (e.g., $\sim$18 steps) than continuous diffusion models~\cite{maskgit}.

While continuous diffusion models utilize guidance sampling to enhance generation quality, MGMs typically utilize sampling with low-temperature Gumbel noise in categorical distribution to enhance the quality ~\cite{maskgit,tokencritic}. However, low-temperature sampling may impose \textit{multi-modality problem}~\cite{gu2017non,ott2018analyzing,zhang2022study}, where the non-autoregressive sampling process fails to generate plausible outputs due to the lack of sequential dependencies. 
As a result, the upper bound of the sample quality generated by MGMs was relatively limited.
Several approaches have been suggested to improve the sampling quality of MGMs, such as discrete predictor-corrector-based methods~\cite{tokencritic,dpc} that train a second transformer to discern the unrealistic tokens.
However, despite the improved sampling, MGMs still underperform in terms of FID scores~\cite{heusel2017gans} compared to state-of-the-art continuous diffusion models such as LDM~\cite{ldm} on ImageNet benchmark~\cite{deng2009imagenet}, even when the model sizes are similar.

\begin{figure}[!t]
  \centering
\includegraphics[width=.9\linewidth]{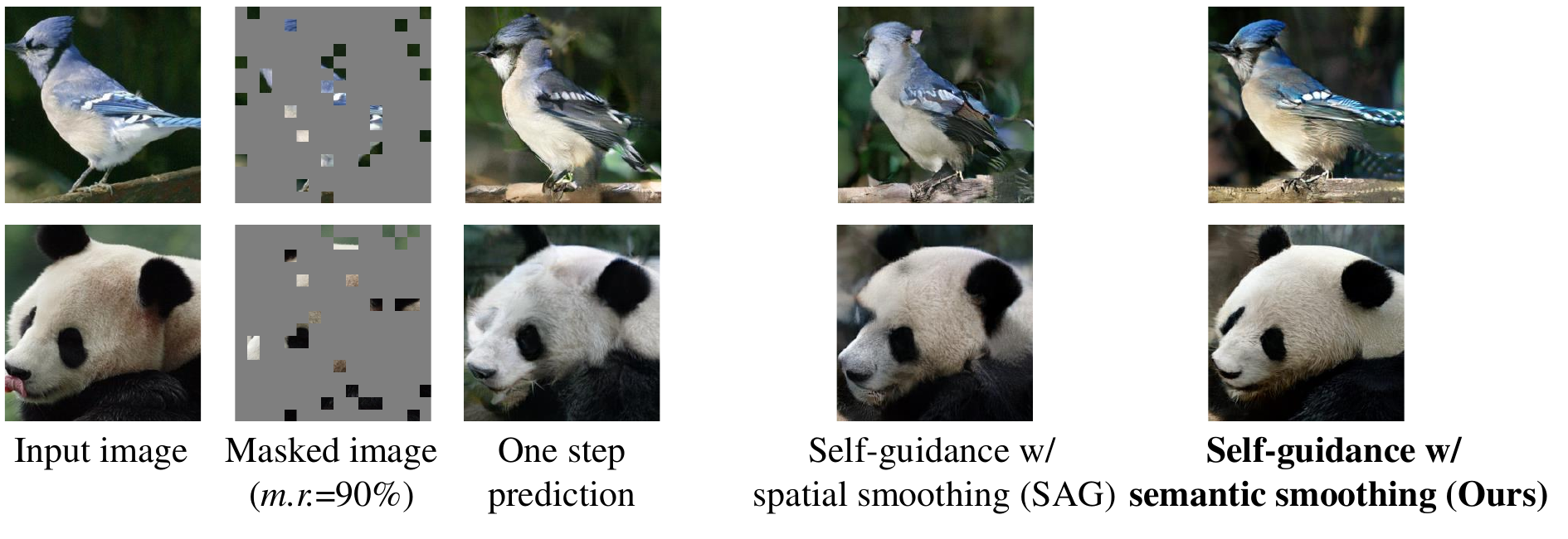}
  \caption{
    Visualization of the effect of guidance using spatial smoothing (SAG)~\cite{sag} and the proposed semantic smoothing. We tokenize the input image using VQGAN~\cite{esser2021taming} encoder, mask the 90\% of VQ tokens, and predict $\hat \vx_{0,t}$ using MaskGIT~\cite{maskgit}. With the proposed self-guidance leveraging semantic smoothing, generated sample quality is improved by enhancing fine-scale details.
  }
  \label{fig:demon}
\end{figure}

To overcome such limitations and enhance the sample quality of MGMs, we propose discrete self-guidance sampling (\cref{fig:visualize}). 
Self-guidance ~\cite{sag} in continuous diffusion models improves the generation quality by enhancing the fine-grained detail of the sample by guiding the diffusion process with coarse-grained information within intermediate diffusion steps, similar to that of classifier-free guidance, which utilizes unconditional generation to enhance the quality of class conditional generation. 
In continuous space, the coarse-grained information can be easily obtained with spatial smoothing, like Gaussian blur. 
To apply self-guidance in MGMs, we first define the general guidance formulation for the discrete diffusion models and introduce self-guidance in MGMs. 
However, while it is simple to define the coarse-grained information in continuous space, the VQ token space~\cite{vqvae2,esser2021taming} in MGMs is absent of continuous semantic structure and cannot apply such a simple strategy, e.g., blur~\cite{berthelot2018understanding}. 
Furthermore, we cannot directly define coarse-grained, i.e., semantically smoothed outputs in the VQ token space. Therefore, we introduce an auxiliary task specifically designed for semantic smoothing of the VQ token space, which enables the network to selectively remove details such as local patterns while preserving overall information in VQ token spaces. 
Consequently, by guiding the MGMs with semantically smoothed information, we can enhance the quality of MGMs (\cref{fig:demon}) even with high temperatures, therefore achieving both high quality and diversity compared to previous MGMs.
For efficient implementation of the discrete self-guidance, we introduce a plug-and-play module with parameter-efficient fine-tuning~\cite{shi2023toast} to pre-trained MGMs, leveraging the generative prior in the MGMs while enabling efficient training with a few parameters and epochs.

Experimental results demonstrate that the proposed guidance effectively improves the sample quality with only 10 epochs of fine-tuning.
Notably, combined with a high sampling temperature, the proposed guidance not only improves the quality of generated samples but also keeps diversity high, showing a superior quality-diversity trade-off compared to other improved sampling techniques for MGMs and other generative families with similar model sizes.

\section{Background}

\subsection{Masked Generative Models}


Discrete diffusion models~\cite{d3pm} aim to generate categorical data $\vx_0 \in \R^{N}$ with a length of $N$ and $K$ categories. The forward Markov process $q(\vx_{t+1}|\vx_t)$ gradually corrupts data $\vx_0,\vx_1,...,\vx_T$ until the marginal distribution of $\vx_T \sim p(\vx_T)$ becomes stationary. Then starting from the $\vx_T$, the learned reverse Markov process $p_\theta(\vx_{t-1}|\vx_t)$ gradually recovers the corrupted data to generate $\vx_0$.

Masked Generative Models (MGMs) are a family of discrete diffusion models that use an \textit{absorbing state diffusion process}~\cite{d3pm},
in which they gradually mask the input tokens by replacing them with a mask token (\texttt{[MASK]}) and learn to predict the masked region. 
Recently, vector-quantized (VQ) image token-based MGMs~\cite{maskgit,tokencritic,dpc} have achieved a superior trade-off between sampling time and quality for image synthesis compared to Gaussian (continuous) diffusion models~\cite{adm,ldm} through the non-autoregressive parallel decoding process.
MGMs are trained to predict the masked region similar to BERT~\cite{devlin2018bert} in natural language processing (NLP), but with various masking ratios $\gamma_t \in (0,1]$, where $\gamma$ is pre-defined mask scheduling function that masks $n=[\gamma(t/T) \cdot N]$ tokens from $N$ total tokens. 
Without loss of generality, we assume the unmasked region of the mask $\vm_t$ is 1, and the masked region is 0.
Then, given external condition $c$ (e.g., class or text) and masked input $\vx_t=\vx_0 \odot \vm_t$, where the mask $\vm_t$ is randomly sampled according to masking ratio $\gamma_t$, MGMs are trained to predict clean data $\vx_0$ (or equivalently, to predict the masked regions) by the objective function
\begin{equation}
    \gL_{mask} = -{\E}_{\vx,t} \left[ \sum_{\forall i \in [1,N], \vm_i=0} \log{p_\theta(\vx_0^i|\vx_t,c)} \right].
    \label{eq:loss_gen}
\end{equation}
Here, we denote the $i$-th token of $\vx_0$ as $\vx_0^i$.


After the training, to sample the image $\vx_0$, MGMs typically use an iterative prediction-masking procedure starting from the blank canvas $\vx_T$ (i.e., all tokens are masked).
Given (partially) masked image token $\vx_t$ sampled from the previous step, MGMs first predict all tokens $\hat{\vx}_{0,t} \sim p_\theta(\hat{\vx}_{0,t}|\vx_t)$ simultaneously, where $\hat{\vx}_{0,t}$ denotes the prediction of $\vx_0$ at timestep $t$ and we omit $c$ for simplification. 
Then $\hat{\vx}_{0,t}$ is masked according to masking ratio $\gamma_{t-1}$ to obtain $\vx_{t-1}$.
MGMs usually sample $\vm_{t-1} \sim p(\vm_{t-1}|\vm_{t},\hat{\vx}_{0,t})$ which is equivalent to sample $\vx_{t-1}$ since $\vx_{t-1}=\vm_{t-1} \odot \hat{\vx}_{0,t}$.
Randomly sampling the $\vm_{t-1}$ can be one choice; however, the prediction $\hat{\vx}_{0,t}$ may have numerous errors, thus randomly selecting the mask can produce sub-optimal results by potentially masking relatively accurate tokens while leaving unrealistic tokens unmasked.
To overcome this, various research adopts improved sampling methods such as utilizing the output confidence of the $\hat \vx_{0,t}$ since confident tokens tend to be more accurate~\cite{maskgit} or train an external corrector to classify the realistic tokens~\cite{tokencritic,dpc}.

However, these improved sampling may impose a problem, named \textit{multi-modality problem} that is a well-known problem in non-autoregressive parallel sampling~\cite{gu2017non,ott2018analyzing,zhang2022study}. Given the input such as $\vx_t$ and class condition $c$, the model can have multiple plausible outputs, which brings challenges to the non-autoregressive model as they generate each token independently. 
In an extreme case where the input token is all masked, and the model predicts output only using a given external condition such as class, each token can predict \textit{easy} token more confidently and correctly, such as a background in every token. 
As a result, correctness-based sampling may result in images only filled with background images.
To resolve this problem, various MGMs adopt additional randomness to sample $\vm_t$, such as sampling with temperature~\cite{maskgit,tokencritic,dpc}. 
Let the $l_t$ measure the realism of sampled tokens $\hat \vx_{0,t}$, such as the confidence scores of sampled tokens in MaskGIT~\cite{maskgit}.
Then MGMs sample $\vm_{t-1}$ by selecting top-$k$ elements of $\tilde l_t = l_t + \tau \cdot (t/T)\vn$ according to $\gamma_t$, where $\vn$ denotes the sampling noise such as i.i.d. Gumbel noise and $\tau$ is temperature scale that is annealed according to the timesteps.
Generally, high-temperature sampling results in more diverse samples while degrading the sample quality.




\subsection{Sampling Guidance}

Iterative sampling processes of diffusion models are often guided by external networks or themselves. Therein, salient information-based guidance has been actively explored in continuous diffusion models~\cite{ho2022classifier,sag,improvedvq} for high-quality image synthesis.
Let $h_t$ be salient information of $\vx_t$ and $\Bar{\vx}_t$ be a perturbed sample that lacks $h_t$. $h_t$ can be internal information within $\vx_t$ or an external condition, or both. 
Lee et al. ~\cite{sag} proposed a general guidance technique that guides the sampling process toward enhancing the information $h_t$, and the equation for the sampling process is:
\begin{equation}
    \tilde{\epsilon}(\bar{\vx}_t, h_t) = \epsilon(\bar{\vx}_t) + (1+s)(\epsilon(\bar{\vx}_t,h_t)-\epsilon(\bar{\vx}_t)),
\label{eq:sag}
\end{equation}
where $\epsilon$ is a score function, $\tilde \epsilon$ is a guided score function of the continuous diffusion model, and $s$ is a guidance scale.
For instance, in the setting $h_t=c$ and $\Bar{\vx}_t=\vx_t$, the \cref{eq:sag}
collapses to classifier-free guidance (CFG)~\cite{ho2022classifier}, which guides the sampling toward the given class distribution. Lee et al.~\cite{sag} propose using adversarial blurring, enhancing the fine-scale details of the sample. 
Generally, using a large guidance scale $s$ enhances the quality of generated samples while reducing the diversity.

Recently, discrete CFG~\cite{improvedvq,chang2023muse} has been introduced to improve the correlation between the input class or text condition $c$ and the images generated by discrete diffusion models as below equation: 
\begin{equation}
    \log p({\tilde \vx}_t) = \log p({\vx}_t) + (1+s)(\log p({\vx}_t|c) - \log p({\vx}_t)),
    \label{eq:dcfg}
\end{equation}
where $\tilde \vx_t$ denotes the guided token.

Unlike CFG in the continuous domain, discrete CFG estimates the probability distribution $p(\vx_t|c)$ directly. However, it requires a specific training strategy and paired labels such as class and text.

\subsection{Parameter-Efficient Fine-Tuning}
MGMs across various domains adopt transformer architecture due to their superior ability to handle context with bidirectional attention~\cite{devlin2018bert,ghazvininejad2019mask,maskgit,yu2023magvit,ziv2024masked}.
However, training a transformer from scratch requires significant computational resources due to the quadratic complexity of the attention mechanism. 
In recent years, parameter-efficient fine-tuning (PEFT) techniques have received significant attention, especially in light of the growing size and complexity of pre-trained models. 
PEFT adapts large pre-trained models to specific tasks or datasets by tuning a small portion of parameters~\cite{zaken2021bitfit,xie2023difffit} or introducing task-specific parameters~\cite{ryu2023low,houlsby2019parameter,shi2023toast}, effectively transferring knowledge without extensive retraining.
Notably, by preserving most of the parameters, the fine-tuned model effectively preserves knowledge with few forgetting.

\section{Methods}

\subsection{Generalized Information-Based Guidance for Discrete Diffusion Models}
Similar to the general guidance in continuous diffusion models in \cref{eq:sag}, discrete CFG in \cref{eq:dcfg} can be extended to the generalized information-based guidance from the optimization perspective.
Given some salient information $h_t$, we aim to sample $\vx_t$ which maximizes $p(\Bar{\vx}_t|h_t)$.
Simultaneously, for the correlation between the information and sample, $p(h_t|\Bar{\vx}_t)$ also needs to be maximized as stated in the equation below which is from Tang et al.~\cite{improvedvq}:
\begin{equation}
\argmax_{\bar\vx_t}[\log p(\Bar{\vx}_t|h_t) + s \log p(h_t|\Bar{\vx}_t)].
\end{equation}
Using the Bayes' theorem and ignoring the prior probability term for salient information, the optimization goal can be represented as:
\begin{equation}
    \argmax_{\bar\vx_t}[\log p(\Bar{\vx}_t) + (1+s)(\log p(\Bar{\vx}_t|h_t) - \log p(\Bar{\vx}_t))].
    \label{eq:dsag}
\end{equation}

Then, the \cref{eq:dsag} guides the sampling process toward enhancing the relevance of the sample and salient information $h_t$.

In the inference stage, various MGMs adopt to predict unmasked state $\hat \vx_{0,t}$ rather than directly predicting $\vx_{t-1}$. 
If we limit $h_t$ to the internal information of $\vx_t$ to make $h_t$ removable from the $\vx_t$ through an information bottleneck module $\gH_\phi$; in other words, if $\Bar \vx_{t}=\gH_\phi({\vx}_{t})$ and $p(\vx_t) = p(\bar{\vx}_t,h_t)$ get satisfied, we can sample next state for the denoising step as below:
\begin{equation}
    \log p_\theta(\tilde {\vx}_{0,t}|{\vx}_{t}) = \log p_\theta(\bar{\vx}_{0,t}|\gH_\phi({\vx}_{t}))) + (1+s)(\log p_\theta({\hat \vx}_{0,t}|{\vx}_{t}) - \log p_\theta(\bar\vx_{0,t}|\gH_\phi({\vx}_{t}))),\\
    \label{eq:dsag_prac}
\end{equation}
when a MGM with parameter $\theta$ predict $\hat\vx_{0,t}$ from ${\vx}_{t}$ and $\bar\vx_{0,t}$ from $\gH_\phi({\vx}_{t})$.
This implies that by defining the information bottleneck module $\gH_\phi$ that can selectively subtract salient information $h_t$ from $\vx_t$ in the discrete space, we can guide the sampling of MGMs in a direction that enhances $h_t$.
Since we aim to improve the sample quality of MGMs by presenting a novel guidance method, it is necessary to define $\gH_\phi$ that can remove information about the fine details of the samples.
For continuous diffusion models, utilizing samples spatially smoothed by Gaussian blur for guidance has been helpful in improving sample quality by restricting fine-scale information~\cite{sag}.
This motivates us to investigate guidance with smoothed output for discrete domains, especially for VQ tokens.

\subsection{Auxiliary Task Learning for Semantic Smoothing on VQ Token space} \label{sec:method:body}
We aim to apply smoothing, such as Gaussian blur, for VQ tokens, as discussed in the previous section, to implement guidance in the discrete domain.
However, unlike natural images which have inherent, observable patterns and structures, the latent space of autoencoders often lacks such properties~\cite{berthelot2018understanding}.
For instance, two successive tokens in VQ codebooks, such as the 11th and 12th tokens, are not semantically linked. As a result, applying Gaussian blur in the VQ token cannot produce meaningful representations. 
Nevertheless, we empirically found that applying Gaussian blur in the probability spaces, i.e. blurring the output logits of the generator $p_\theta(\hat \vx_0|\vx_t)$ similar to Lee et al.~\cite{sag} can provide meaningful guidance.
However, the improvement is marginal because Gaussian blur is not a suitable information bottleneck for subtracting fine details in VQ token space.

To overcome this, we introduce an auxiliary task designed to leverage semantic smoothing for VQ tokens, a process that selectively removes details such as local patterns while preserving overall information in VQ token space.
Specifically, given the masked input $\vx_t$, masked generator predicts $p_\theta(\hat \vx_{0,t}|\vx_t)$.
We aim to train information bottleneck $\gH_\phi$ to generate the semantically smoothed output $p_\theta(\hat \vx_{0,t}|\gH_\phi (\vx_{t}))$.
%
However, training $\gH_\phi$ directly is challenging, as we cannot define semantically smoothed outputs.
To naturally impose a model to generate semantically smoothed output, we leverage error token correction, originally introduced in non-autoregressive machine translation to mitigate the compounding decoding error in the iterative sampling process~\cite{cmlmc}.
During the error token correction, input unmasked tokens are randomly replaced with error tokens, and the model learns to correct them. 
To be more specific, let $\vz_t$ be a corrupted data where some tokens in $\vx_t$ are replaced with error tokens with some probability $p$.
Then, the objective function for the auxiliary task to update $\phi$ can be
\begin{equation}
    \gL_{aux} = -{\E}_{\vx, t} \left[ \sum_{\forall i \in [1,N], \vm_i=0} \log{p_\theta(\vx_0^i|\gH_\phi(\vz_t),c)} \right].
    \label{eq:loss_corr}
\end{equation}
From the perspective of \textit{Vicinal Risk Minimization} (VRM)~\cite{chapelle2000vicinal,zhang2017mixup}, a vicinity distribution $p_\phi$ minimizes the empirical risk for all data points $\vx_t$ given a vicinity of the data $\vz_t$.
Given that randomly replaced error tokens often act as semantic vicinities within the input data, to minimize the overall empirical risk, the model implicitly learns to smooth vicinities of $\vz_t$. 
This involves leveraging coarse information from the surrounding context while minimizing the fine-scale details in the presence of unknown input errors\footnote{For a simple example, a common approach to correcting unknown \textit{numerical outliers} in a 1-dimensional signal, such as impulse noise in time series data, is smoothing the signal, like applying low-pass filters. Similarly, to correct the unknown error tokens, which are \textit{semantic outliers} in our case, we expect that the model implicitly learns to smooth $\vz_t$ to deal with unknown error tokens.}.
However, training a network from scratch with \cref{eq:loss_corr} does not ensure that the outputs will resemble real images, potentially converging on trivial solutions that may be undesirable, in addition to being computationally expensive.


\begin{figure}[!t]
  \centering
\includegraphics[width=1.0\linewidth]{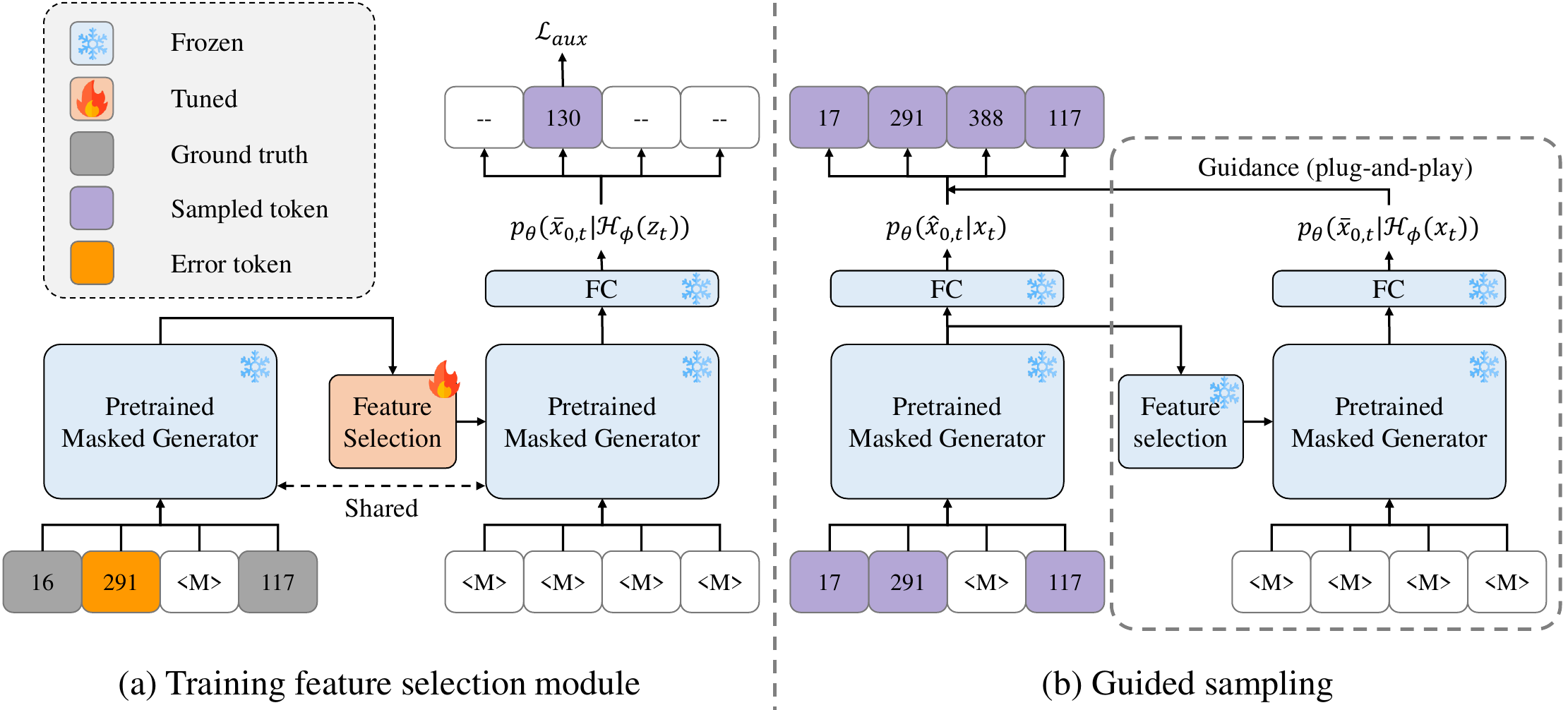}
  \caption{(a) Fine-tuning the feature selection module $\gH_\phi$ (TOAST~\cite{shi2023toast}). With the auxiliary objective in \cref{eq:loss_corr}$, \gH_\phi$ implicitly learns to smooth erroneous input $\vz_t$ to address semantic outliers (\cref{sec:method:body}).
  (b) During the sampling steps, self-guidance can be efficiently implemented by leveraging the feature map from the generative process. 
  $\gH_\phi$ performs semantic smoothing on the input $\vx_t$, guiding the sampling process toward enhancing fine-scale details in the generated sample.
  }
  \label{fig:method}
\end{figure}

\subsection{Efficient Implementation} \label{sec:method:toast}
To mitigate the aforementioned problem, we adopt a parameter-efficient fine-tuning (PEFT) method to utilize deep image priors in the pre-trained masked generator and to enhance the training efficiency of transfer learning.
Among various PEFT methods, we adopt TOAST~\cite{shi2023toast}, which shows favorable performance in various visual and linguistic tasks under transformer architecture.
With a frozen pre-trained backbone, TOAST selects task-relevant features from the output and feeds them into the model by adding them to the value matrix of self-attention. 
This top-down signal steers the attention to focus on the task-relevant features, effectively transferring the model to other tasks without changing parameters.

Besides its effectiveness in transfer learning, TOAST brings more practical strengths to our task. 
Before the discussion, it is important to note that masked generative models exhibit a strong training bias. Because the training data does not contain error tokens, the model predicts unmasked input as identical and unable to correct error tokens.
Since the model is trained only to consider masked regions, we propose to use a blank canvas as an input (i.e., all tokens are masked $\vx_T$), enabling the model to make corrections in response to error tokens. However, most PEFT methods lack a direct solution for incorporating information about $\vx_t$ when the input is replaced with all masked tokens.
On the other hand, we empirically found that simply replacing the input for the second stage of TOAST as $\vx_T$ can mitigate this problem without a performance drop (\cref{fig:method} a).

Furthermore, the two-stage approach of TOAST can be efficiently implemented in the sampling process.
The generator first sample $p_\theta(\hat \vx_{0,t}|\vx_t)$ from the input $\vx_t$. 
The hidden state obtained from the generation stage can be recycled for the second stage to produce guidance logit $p_\theta(\bar \vx_{0,t}|\gH_\phi(\vx_t))$ (\cref{fig:method} b).
We note that different from the fine-tuning step where the error tokens $\vz_t$ are used to train $\gH_\phi$, the sampling process directly utilizes $\vx_t$ to generate the semantically smoothed tokens $\gH_\phi(\vx_t)$.

\section{Related Works}

\textbf{Generative Adversarial Networks (GANs).} Trained by adversarial objective~\cite{GAN}, GANs have shown impressive performance in image synthesis with only 1-step model inference~\cite{karras2019style,karras2020training,stylegan-xl,biggan,lee2023fix}. However, despite their practical performance, the training instability and limited mode coverage caused by the adversarial objective~\cite{salimans2016improved, zhao2018bias} is still a bottleneck for their broader applications.

\textbf{Diffusion Models.}
Recent diffusion models in continuous space mostly utilize Gaussian noise to perturb the input data and learn to predict clean data from the noisy data~\cite{sohl2015deep}. 
After the advent of DDPM~\cite{ho2020denoising}, diffusion models have rapidly grown with architecture improvements~\cite{nichol2021improved,dit}, improved training strategies~\cite{kingma2021variational,choi2022perception,hur2024expanding}, improved samplings~\cite{DDIM,lu2022dpm}, latent space models~\cite{ldm}, and sampling guidance~\cite{adm,ho2022classifier,sag}, outperforming various generative families such as GANs in various domains. 

\paragraph{Masked Generative Models.}
With the recent success of transformers~\cite{vaswani2017attention} and GPT~\cite{radford2018gpt} in NLP, generative transformers have also been adopted in image synthesis. To mimic the generative process of transformers for discrete embedded natural language, recent studies encode images into quantized visual tokens using the VQ-VAE encoder \cite{vqvae2,esser2021taming} and apply the generation process in an autoregressive \cite{lee2022autoregressive, vit-vqgan} or non-autoregressive manner \cite{maskgit,tokencritic,dpc,chang2023muse}. 
In particular, non-autoregressive generation shows a better trade-off between generation quality and speed. 
Nevertheless, they suffer from \textit{compounding decoding errors} \cite{dpc}, which means that small decoding errors in early generation steps can accumulate into large differences in later steps. 
To address this issue, Token-Critic \cite{tokencritic} and DPC \cite{dpc} use additional transformers to identify more realistic tokens. 
However, they require a training second transformer, which incurs a training cost similar to training a generator from scratch.

The non-autoregressive predict-mask sampling of MGMs can be regarded as a discrete diffusion process that uses \textit{absorbing state diffusion process}~\cite{d3pm,dpc}. 
Similar to MGMs, VQ-Diffusion~\cite{vqdiffusion} proposed a mask-and-replace diffusion strategy to predict masked tokens.
Improved VQ-Diffusion~\cite{improvedvq} adopt purity-based sampling with discrete CFG to further improve the sample quality.

\begin{table}[!t]
\centering
\small
{
\setlength{\tabcolsep}{4pt}
\caption{
{Quantitative comparison of various generative models for class-conditional image generation on ImageNet 256$\times$256 and 512$\times$512 resolutions}.
``$\downarrow$'' or ``$\uparrow$'' indicate lower or higher values are better.
$\dag$: taken from MaskGIT~\cite{maskgit}, $\ddagger$: taken from VAR~\cite{var}, $\ast$: taken from Token-Critic~\cite{tokencritic}.
}\label{tab:main}
\scalebox{0.93}
{
\begin{tabular}{lccccccccccc}
\toprule
 &&& \multicolumn{4}{c}{ImageNet 256$\times$256} && \multicolumn{4}{c}{ImageNet 512$\times$512}\\ \noalign{\vskip 1pt}
\cline{4-7} \cline{9-12} \noalign{\vskip 1pt}
Model & Type  & NFE & FID$\downarrow$ & IS$\uparrow$ & Prec$\uparrow$ & Rec$\uparrow$ && FID$\downarrow$ & IS$\uparrow$ & Prec$\uparrow$ & Rec$\uparrow$\\
\midrule
BigGAN-deep~\cite{biggan} & GANs & 1  & 6.95  & 224.5   & \textbf{0.89} & 0.38 && 8.43 & 177.9 & 0.85 & 0.25   \\
GigaGAN~\cite{gigagan}   & GANs & 1    & 3.45  & 225.5       & 0.84 & 0.61 && $-$ & $-$ & $-$ & $-$   \\
\midrule
ADM~\cite{adm} & {Diff.}  & 250      & 10.94 & 101.0      & 0.69 & \textbf{0.63} && 23.24 & 58.0 & 0.73 & \textbf{0.60}  \\
ADM (+ SAG)~\cite{sag} & {Diff.}  & 500    & 9.41 & 104.7        & 0.70 & 0.62  && $-$ & $-$ & $-$ & $-$  \\
CDM~\cite{cdm} & {Diff.} & 250   & 4.88  & 158.7   & $-$  & $-$ && $-$ & $-$ & $-$ & $-$   \\
LDM-4~\cite{ldm} & {Diff.} & 250 & 10.56  & 103.4  & 0.71  & 0.62 &&  $-$ & $-$ & $-$ & $-$   \\
LDM-4 (+ CFG)~\cite{ldm} & {Diff.}  & 500  & 3.60  & {247.7}       & $-$  & $-$  &&  $-$ & $-$ & $-$ & $-$  \\
DiT-L/2$^\ddagger$ (+ CFG)~\cite{dit} & {Diff.}  & 500  & 5.02  & 167.2       & 0.75 & 0.57 && $-$ & $-$ & $-$ & $-$ \\
\midrule
VQVAE-2$^\dag$~\cite{vqvae2}  & {AR} & 5120  & 31.11      & $\sim$45     & 0.36           & 0.57  &&   $-$ & $-$ & $-$ & $-$     \\
VQGAN$^\dag$~\cite{esser2021taming} & AR & $\sim$1024  & 18.65 & 80.4         & 0.78 & 0.26  && 7.32 & 66.8 & 0.73 & 0.31\\
\midrule
VQ-Diffusion~\cite{vqdiffusion} & {Discrete.} & 100 & 11.89 & $-$ & $-$ & $-$ && $-$ & $-$  & $-$ & $-$ \\
ImprovedVQ. (+ CFG)~\cite{improvedvq} & {Discrete}. & 200 & 4.83 & $-$ & $-$ & $-$ && $-$ & $-$  & $-$ & $-$ \\
\midrule
MaskGIT$^\ast$~\cite{maskgit} & Mask. & 18  & 6.56  & 203.6        & 0.79 & 0.48  && 8.48 & 167.1 & 0.78 & 0.46   \\
Token-Critic~\cite{tokencritic} & Mask. & 36  & 4.69  & 174.5        & 0.76 & 0.53 && 6.80 & 182.1 & 0.73 & 0.50   \\
DPC-light~\cite{dpc} & Mask.  & 66  & 4.8  & 249.0       & 0.80 & 0.50  && 6.09 & 228.1 & 0.81 & 0.46  \\
DPC-full~\cite{dpc} & Mask.  & 180  & 4.45  & 244.8        & 0.78 & 0.52  && 6.06 & 218.9 & 0.80 & 0.47  \\
\midrule
Ours (T=12) & Mask.  & 24 & 3.35  & 259.7        & 0.81 & 0.52 && \textbf{5.38} & 226.0 & \textbf{0.88} & 0.36    \\
Ours (T=18) & Mask.  & 36 & \textbf{3.22}  & \textbf{263.9}        & 0.82 & 0.51 && {5.57} & \textbf{233.2} & \textbf{0.88} & 0.35    \\
\bottomrule
\end{tabular}
}
}
\end{table}

\section{Experiments}


\textbf{Datasets, Baselines, and Metrics.} We demonstrate the effectiveness of the proposed guidance for masked generative models on class conditional generation using the Imagenet benchmark~\cite{deng2009imagenet} with $256\times256$ and $512\times512$ resolutions.  For a baseline model, we use MaskGIT~\cite{maskgit}, which shows a state-of-the-art trade-off between quality and sampling speed on a class conditional generation of MGMs and has publicly available checkpoints for our target datasets.
To evaluate the trade-off between sample fidelity and diversity, we measure Fr\'echet Inception Distance (FID)~\cite{heusel2017gans}, Inception Score (IS)~\cite{salimans2016improved}, Precision, and Recall~\cite{kynkaanniemi2019improved} using the implementation provided by Dhariwal et al.~\cite{adm}. 
To measure the computational cost, we report the number of function evaluations (NFE) required to sample an image. 
Note that total sampling timesteps ($T$) may differ from NFE due to guidance.

\textbf{Implementation Details.}
We use a VQGAN tokenizer~\cite{esser2021taming} provided by MaskGIT~\cite{maskgit}, which encodes images into 10-bit integers.
Since the training code for MaskGIT is unavailable, we implement based on the open Pytorch reproduction~\cite{open-muse}.
We follow the previous work for error token correction~\cite{cmlmc} to prepare input with error tokens and randomly replace 30\% of input tokens with error tokens.
We utilized an NVIDIA RTX A6000 for fine-tuning and sampling. We used an exponential moving average (EMA) of fine-tuning weights with a decay of 0.9999 and \texttt{bf16} precision. The batch size was set to 256, and the additional parameters introduced by TOAST are approximately 20-25\% of the model size. Notably, the fine-tuning was completed efficiently within 10 epochs. More detailed implementation of TOAST~\cite{shi2023toast} is provided in the \cref{appendix:impl}.
We use sampling step $T=18$ and sampling temperature $25$ for Imagenet 256$\times$256 and 45 for Imagenet 512$\times$512.
For sampling step $T=12$, we use temperatures 10 and 20, respectively for each resolution.

\begin{figure}[!t]
  \centering
\includegraphics[width=.97\linewidth]{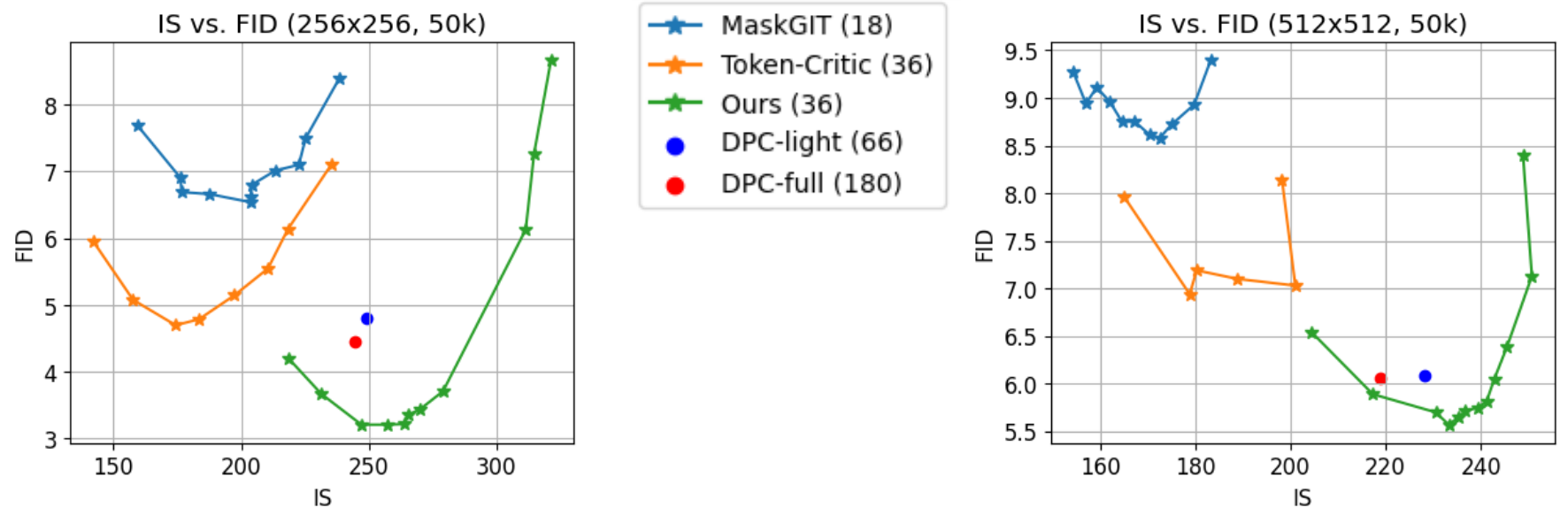}
\vspace{-2mm}
  \caption{
  IS vs. FID curves of various sampling methods for MGMs on ImageNet 256$\times$256 and 512$\times$512.
  The curve positioned towards the bottom right indicates a better trade-off between sample quality and diversity.
  We plot the curve by varying the sampling temperature ($\tau$), and the curves of MaskGIT~\cite{maskgit} and Token-Critic~\cite{tokencritic} are taken from Token-Critic~\cite{tokencritic}.
  }
  \label{fig:tradeoff}
\end{figure}

\subsection{Comparison with Various Generative Models}
\textbf{Quantitiatve results.}
We compare the performance of the proposed method with various class conditional generative methods.
(1) \textit{GANs}: BigGAN~\cite{biggan} and GigaGAN~\cite{gigagan};
(2) \textit{continuous diffusion models} (Diff.): ADM~\cite{adm}, CDM~\cite{cdm}, LDM~\cite{ldm}, and DiT~\cite{dit};
(3) \textit{auto-regressive models} (AR): VQVAE-2~\cite{vqvae2} and VQGAN~\cite{esser2021taming}; 
(4) \textit{discrete diffusion models} (Discrete.): VQ Diffusion~\cite{vqdiffusion}, and Improved VQ Diffusion~\cite{improvedvq};
(5) \textit{masked generative models} (Mask.): MaskGIT~\cite{maskgit}, Token-Critic~\cite{tokencritic}, DPC~\cite{dpc}.
%
As noted in previous literature~\cite{ho2022classifier,dpc}, sampling using a pre-trained classifier may impact the classifier-based metrics such as FID and IS.
Thus, to see the base generative capacity of each method, we compare the models that do not use external pre-trained networks such as a classifier or upsampler during training or sampling.
We also exclude large-scale models such as DiT-XL/2~\cite{dit} for a fair comparison.
%
\cref{tab:main} presents the quantitative results of various generative models with guidance.
All values are taken from the original paper unless otherwise noted.
The proposed method achieves superior FID and IS despite the low computational cost for sampling.
For instance, the proposed method achieves better FID and IS than DiT-L/2 with CFG on Imagenet 256$\times$256 even though the proposed method requires an order of magnitude fewer sampling steps.

We comprehensively compare the proposed methods with various sampling strategies for MGMs in \cref{fig:tradeoff}.
We note that all MGMs use the same VQGAN tokenizer~\cite{esser2021taming} provided in MaskGIT~\cite{maskgit}, the same baseline generator~\cite{maskgit}, and the same sampling timestep $T=18$.
Although the previous methods, such as Token-Critic~\cite{tokencritic} and DPC~\cite{dpc}, require similar or more NFEs to sample images and more training resources to train an external corrector transformer, our simple guidance shows a better trade-off between sample quality and diversity. 

\begin{figure}[!t]
  \centering
\includegraphics[width=.97\linewidth]{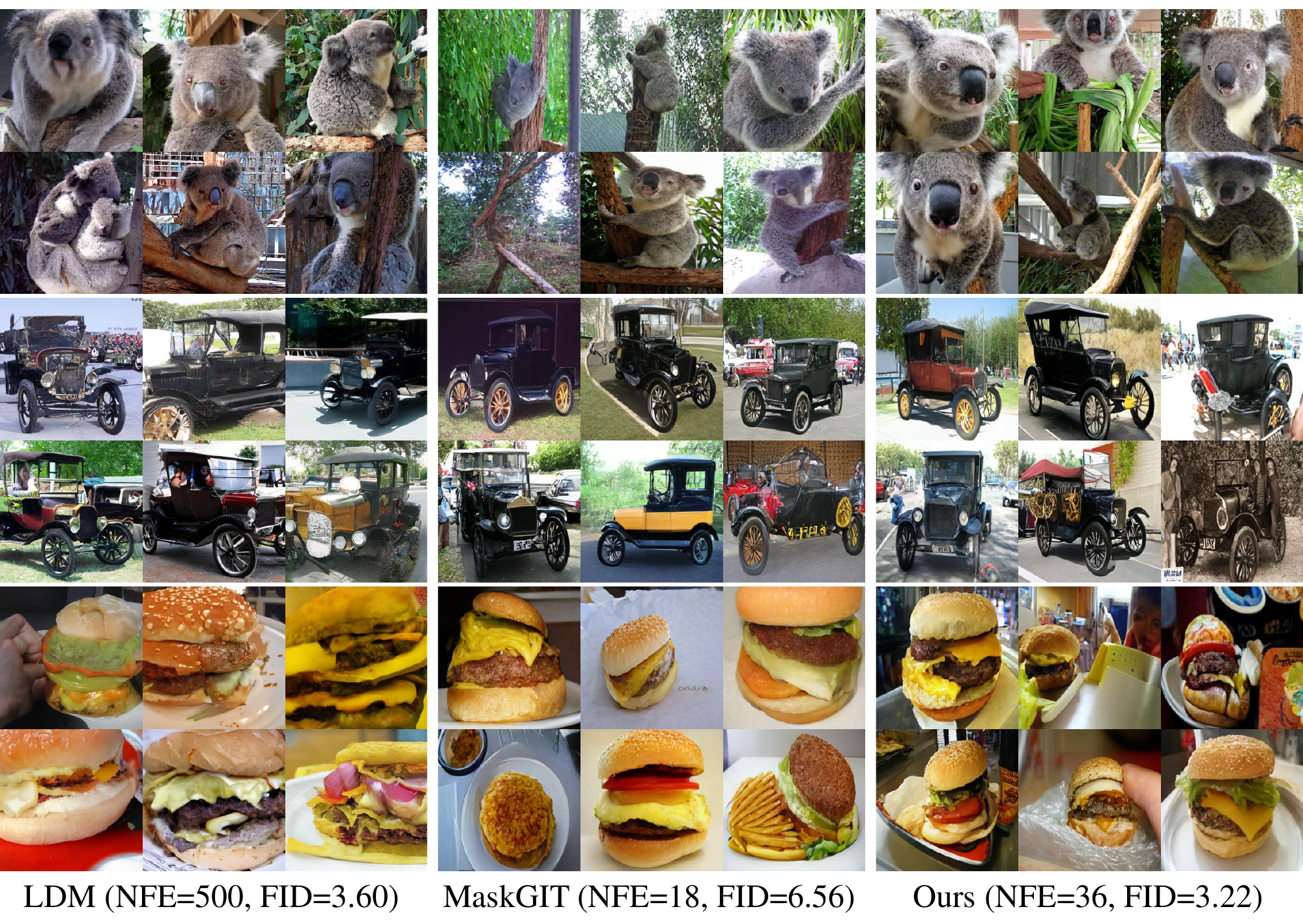}  
  \vspace{-2mm}
  \caption{
  Sampled images on ImageNet 256$\times$256 class conditional generation using selected classes (\texttt{105}: Koala, \texttt{661}: model T, and \texttt{933}: Cheeseburger).
  left: LDM~\cite{ldm} + CFG (s=1.5, NFE=250$\times$2), middle: MaskGIT (NFE=18), right: Ours (s=1.0, NFE= 18$\times$2). 
  }
  \label{fig:qual}
\end{figure}

\textbf{Qualitative results.}
\cref{fig:qual} presents the randomly sampled images using LDM~\cite{ldm} with CFG, MaskGIT, and MaskGIT with the proposed guidance. 
The proposed guidance sampling enhances details in the generated samples while maintaining high diversity. 
More sampled results with various classes are provided in the \cref{fig:visualize} and \cref{appendix:visuazlie}.


\subsection{Ablation Studies and Analysis}

In this section, we explore the effectiveness of the guidance on class conditional ImageNet generation in $256\times256$ scale across the various sampling hyperparameters.
We vary each sampling parameter and otherwise use the default settings.

\begin{wraptable}[11]{r}{0.36\textwidth}
\centering
\small
{
\vspace{-6mm}
\caption{
Ablation results of various guidances on ImageNet 256$\times$256 class conditional generation.
}\label{tab:ablation}
\vspace{2mm}
{
\begin{tabular}{c|cc}
\toprule
 & FID$\downarrow$ & IS$\uparrow$ \\ 
\midrule
Blur Guidance & 8.32 & 231.2 \\
SAG~\cite{sag} & 4.73 & 177.3\\
ft. w/ $\gL_{mask}$ & 4.11 & 238.4 \\
\midrule
ft. w/ $\gL_{aux}$ (Ours) & \textbf{3.22}  & \textbf{263.9} \\ 
\bottomrule
\end{tabular}
}
}
\end{wraptable}

\textbf{Effectiveness of Auxiliary Task.}
To demonstrate that fine-tuning with a proposed auxiliary task using the objective in \cref{eq:loss_corr} can effectively improve the generative capabilities of MGMs, we conduct ablation studies and report the FID and IS in \cref{tab:ablation}. 
As noted in \cref{sec:method:body}, applying Gaussian blur in the input VQ token does not properly smooth VQ tokens. However, we empirically found that applying Gaussian blur in the output logit can produce meaningful guidance. Therefore, we provide the quantitative comparison with guidance by applying Gaussian blur in the logit space (Blur Guidance). We further utilize self-attention value for adversarial masking following Lee et al.~\cite{sag}.
Blur guidance and SAG enhance the quality (IS) or diversity (FID) in some degree, but the improvement is marginal compared to ours.
We further compare the results when the TOAST module is fine-tuned with the generative objective in \cref{eq:loss_gen} to verify the effectiveness of the auxiliary task (ft. w/ $\gL_{mask}$).
Since the TOAST architecture and blank input for the second stage naturally play the role of information bottleneck, the performance has increased.
Nevertheless, fine-tuning with the proposed auxiliary loss in \cref{eq:loss_corr} demonstrates its effectiveness, showing superior performance in both metrics.


\begin{figure}[!t]
  \centering
\includegraphics[width=\linewidth]{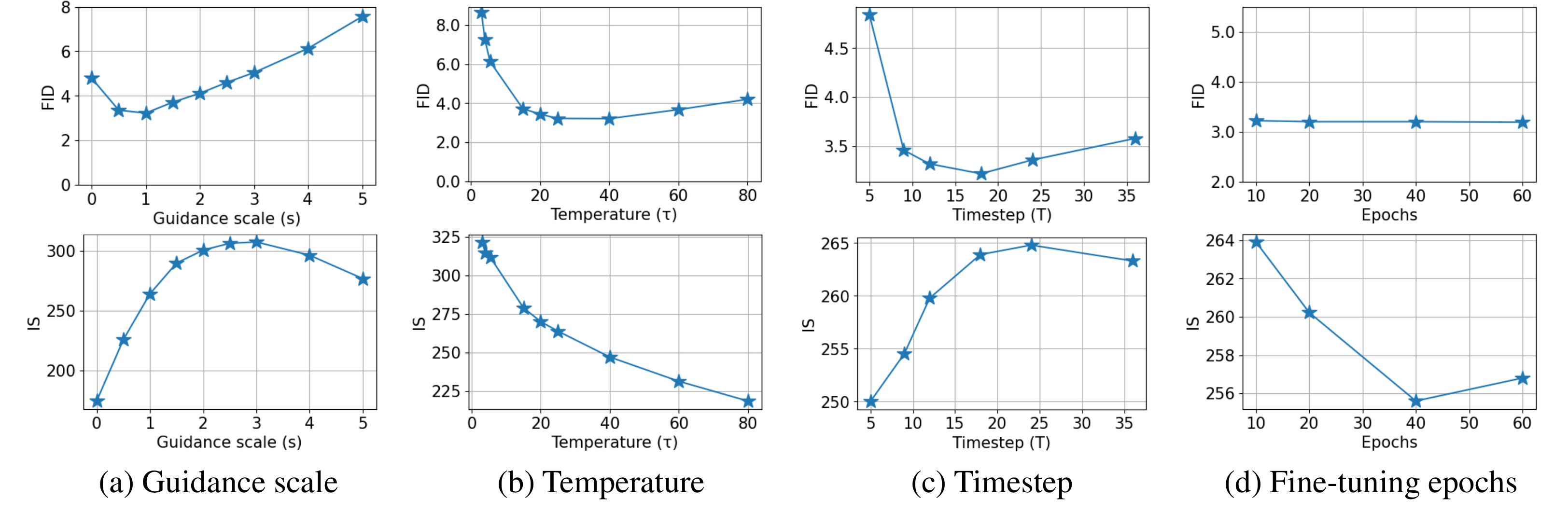}
  \vspace{-5mm}
  \caption{
  Exploring the sampling hyperparameters by varying (a) guidance scale, (b) sampling temperature, (c) sampling timesteps, and (d) fine-tuning epochs.
  }
  \vspace{-2mm}
  \label{fig:abla}
\end{figure}

\textbf{Varying the guidance scale (\cref{fig:abla} a).} 
In line with previous literature on sampling guidance~\cite{adm,ho2022classifier,sag}, a high guidance scale improves the sample quality while sacrificing diversity. We found that the guidance scale 1.0 shows the best performance in terms of FID score and the best trade-offs. However, strong guidance ($s>3$) does not ensure quality improvement, often leading to undesirably highlighted details or saturated colors (\cref{appendix:high_ts}), similar to the effect of large scale with CFG~\cite{ho2022classifier}.

\textbf{Varying the sampling temperature (\cref{fig:abla} b).}
Compared to MaskGIT~\cite{maskgit} limit their sampling temperature relatively low value ($\tau=4.5$), we found that with the proposed guidance, the sample quality can be effectively preserved with a higher sampling temperature. As a result, with a high sampling temperature ($\tau=25$), we achieve better quality and diversity compared to MaskGIT.

\textbf{Varying the sampling steps $T$ (\cref{fig:abla} c).}
We observed that for higher $T$, the optimal sampling temperature also increases. For example, at $T=5$, the best results are achieved with a sampling temperature of 4, while at $T=18$, the optimal sampling temperature rises to 25.
Thus, we plot the best FID and IS for each timestep using different temperatures. 
Whereas the optimal performance-efficiency trade-off of MaskGIT is observed around $T=8$, ours shows such "sweet spot" around $T=18$. Up to this point, both quality and diversity increase as $T$ increases, consistently outperforming MaskGIT.
Furthermore, with fewer NFEs ($T=5$), the proposed method achieves an FID of 4.84 and an IS of 249.9, outperforming the MaskGIT samples with $T=18$.
This demonstrates that the proposed guidance technique provides more scalable and efficient sampling for MGMs.

\textbf{Varying fine-tuning epochs (\cref{fig:abla} d).}
We found that the proposed fine-tuning is efficiently trained within 10 epochs and that no further fine-tuning is required to achieve better results.

\section{Conclusion and Future Work}
In this paper, we define generalized guidance for discrete diffusion models, as a counterpart for guidance in continuous domain~\cite{sag}. To generate guidance, we propose an auxiliary task to apply semantic smoothing in VQ tokens. Experimental results show that the proposed guidance effectively and efficiently improves the generative capabilities of MGMs on class conditional image generation.

\textbf{Future work.} 
Although the proposed guidance can improve the generative capabilities of MGMs on class conditional image generation, there is still room for improvement in various aspects: (1) experiments on large-scale text conditional generative models~\cite{chang2023muse}, (2) generalization to discrete diffusion models~\cite{vqdiffusion}, and (3) generalization to various domains such as audio~\cite{ziv2024masked}, and video~\cite{yu2023magvit}.

\textbf{Societal impacts.} 
While our research does not directly touch ethical issues, the rapid growth of generative models raises considerations in AI ethics (e.g., potential misuse in creating deepfakes~\cite{lee2024selfswapper}, generating antisocial content~\cite{gandikota2023erasing}, or vulnerabilities to adversarial attacks~\cite{zhang2020secret,han2023reinforcement,koh2023disposable}).

\section*{Acknowledgements}
This work was partly supported by Institute of Information \& Communications Technology Planning \& Evaluation(IITP) grant funded by the Korea government(MSIT) (RS-2024-00439020, Developing Sustainable, Real-Time Generative AI for Multimodal Interaction, SW Starlab), Artificial intelligence industrial convergence cluster development project funded by the Ministry of Science and ICT(MSIT, Korea)\&Gwangju Metropolitan City, and the National Research Foundation of Korea(NRF) grant funded by the Korea government(MSIT) (No. RS-2023-00240379).

{\small
\bibliographystyle{abbrvnat}
\bibliography{egbib}
}

\newpage
\appendix
\section{Detailed Implementation of feature selection module (TOAST)}
\label{appendix:impl}

The architecture of the proposed sampling method consists of two parts: MaskGIT~\cite{maskgit} and TOAST~\cite{shi2023toast}.
Since we have not changed the MaskGIT architecture for plug-and-play sampling guidance, we briefly explain the implementation details of TOAST below. 
The TOAST modules consist of three parts: token selection module, channel selection module, and linear feed-forward networks.

(i) The token selection module selects the task or class-relevant tokens by measuring the similarity with the learnable anchor vector $\xi_c$. We generate class conditional anchors $\xi_c$ with simple class conditional MLPs.

(ii) The channel selection is applied with learnable linear transformation matrix $\rmP$. Then, the output of the token and channel selection module is calculated via $\vz_i = \rmP \cdot sim(\vz_i, \xi_c)$, where $\vz_i$ denotes the $i$-th input token.

(iii) After the feature selection, the output is processed with $L$ layer MLP layers, where $L$ is equal to the number of Transformer’s layers. The output of the $l$-th layer of MLP blocks is added to the value matrix of the attention block in $(L-l)$-th Transformer layer (top-down attention steering). Following the previous work~\cite{shi2023toast}, we add variational loss to regularize the top-down feedback path. A more detailed process and theoretical background can be found in previous works on top-down attention steering~\cite{shi2023top,shi2023toast}.

\section{Effect of High Sample Temperature and Guidance Scale}
\label{appendix:high_ts}
\begin{figure}[!h]
  \centering
\includegraphics[width=.5\linewidth]{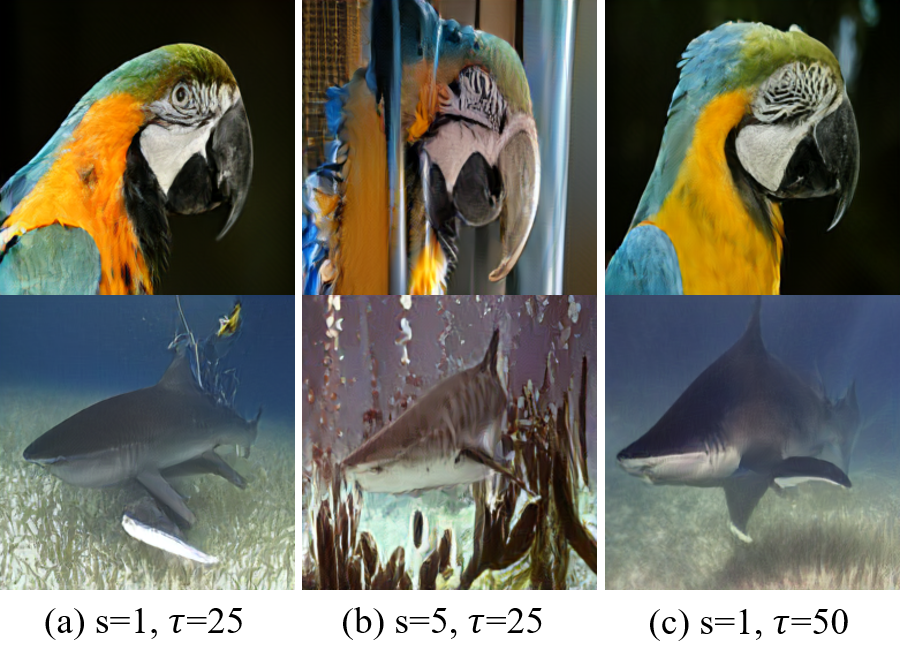}  
  \vspace{-2mm}
  \caption{
  Sampled images on ImageNet 256$\times$256 class conditional generation using (a) our default config, (b) large guidance scale ($s=5$), and (c) high sampling temperature ($\tau=50$). 
  }
  \label{fig:high_ts}
\end{figure}

We show the effect of high sampling temperature ($\tau$) and high guidance scale in \cref{fig:high_ts}. The high sampling temperature (i.e., strong guidance) often leads to undesirably highlighted fine-scale details or saturated colors, similar to the large scale of CFG~\cite{ho2022classifier}. High temperatures often lead to the collapse of the overall structure of generated samples.

\newpage
\section{More Visual Results}
\label{appendix:visuazlie}

\begin{figure}[!h]
  \centering
\includegraphics[width=1.0\linewidth]{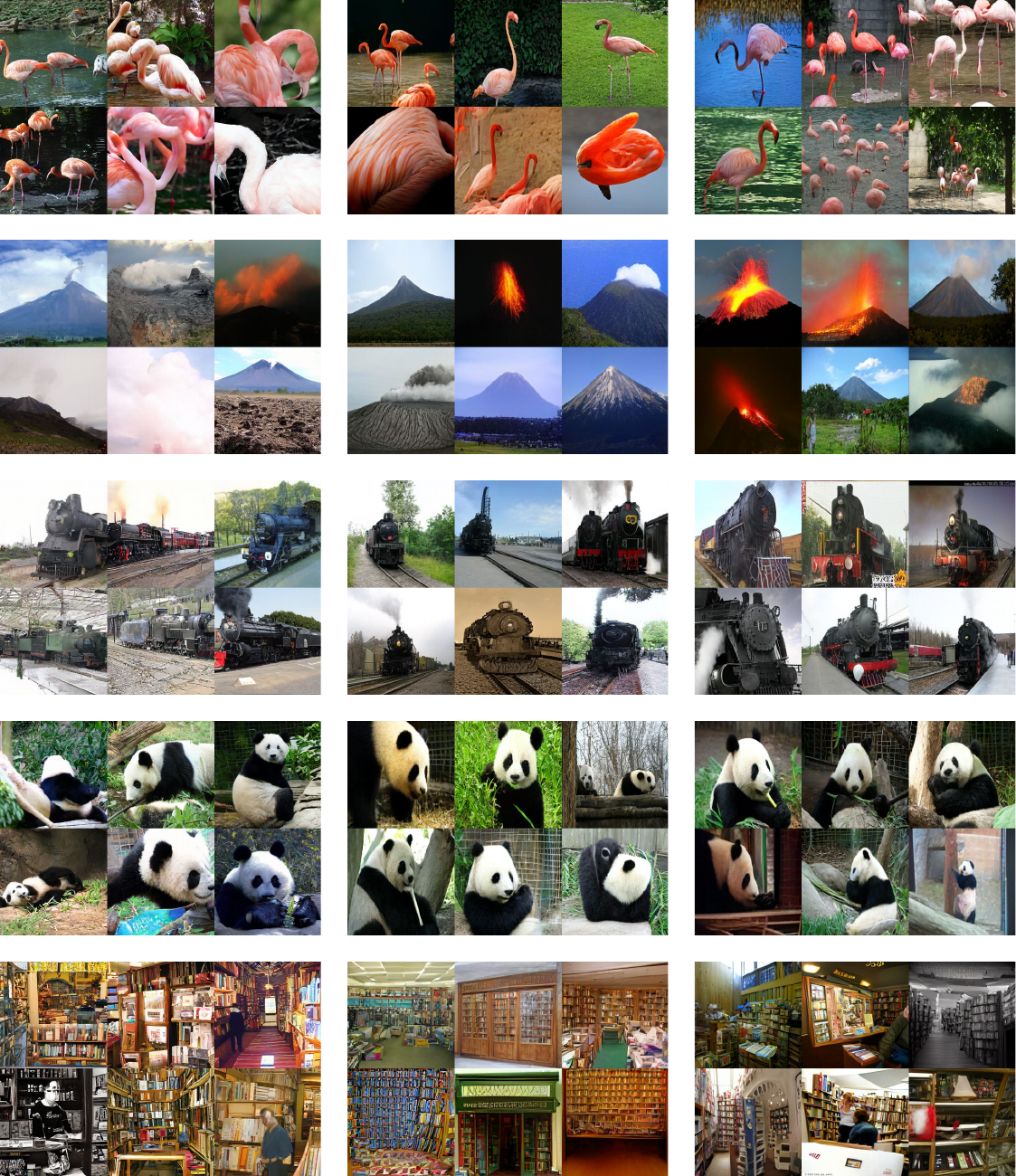}  
  \vspace{-2mm}
  \caption{
  Sampled images on ImageNet 256$\times$256 class conditional generation using selected classes (\texttt{130}: Flamingo, \texttt{980}: Volcano, \texttt{820}: Steam locomotive, \texttt{388}: Giant panda, and \texttt{454}: Bookshop).
  left: LDM~\cite{ldm} + CFG (s=1.5, NFE=250$\times$2), middle: MaskGIT (NFE=18), right: Ours (s=1.0, NFE= 18$\times$2).
  }
\end{figure} 

\end{document}